\pdfoutput=1

\documentclass[11pt]{article}

\usepackage[]{EMNLP2022}

\usepackage{times}
\usepackage{latexsym}

\usepackage[T1]{fontenc}

\usepackage[utf8]{inputenc}

\usepackage{microtype}

\usepackage{inconsolata}

\usepackage{algorithm}

\usepackage{microtype}
\usepackage{boldline}
\usepackage{multirow}
\usepackage{makecell}
\usepackage{hhline}
\usepackage{pgfplots}
\usepackage{tikz}
\usepackage{enumitem}
\usepackage{colortbl}

\usepackage{mathtools}
\usepackage{pifont}

\usepackage{bbm}
\usepackage{amsfonts}
\usepackage{algpseudocode}
\usepackage{booktabs}
\usepackage{xparse}
\usepackage{graphicx}
\usepackage{multicol}
\usetikzlibrary{shapes,arrows}
\usepackage{natbib}
\usepackage{caption}
\usepackage{subcaption}
\usepackage{tikz-dependency}
\usepackage{wrapfig}
\pgfplotsset{compat=1.17} 
\usepackage{subfiles}
\usepackage{readarray}
\usepackage{xcolor}
\usepackage{import}

\newcommand{\mcL}{\mathcal{L}}

\usepackage{amsmath,amsfonts,bm}

\def\eqref#1{equation~\ref{#1}}

\def\1{\bm{1}}

\def\vb{{\bm{b}}}

\def\vk{{\bm{k}}}

\def\vq{{\bm{q}}}
\def\vr{{\bm{r}}}

\def\vt{{\bm{t}}}

\def\vv{{\bm{v}}}

\def\vx{{\bm{x}}}
\def\vy{{\bm{y}}}
\def\vz{{\bm{z}}}

\def\mI{{\bm{I}}}

\def\mM{{\bm{M}}}

\def\mU{{\bm{U}}}

\def\mW{{\bm{W}}}

\DeclareMathAlphabet{\mathsfit}{\encodingdefault}{\sfdefault}{m}{sl}
\SetMathAlphabet{\mathsfit}{bold}{\encodingdefault}{\sfdefault}{bx}{n}

\def\sR{{\mathbb{R}}}

\newcommand{\sigmoid}{\sigma}

\DeclareMathOperator*{\argmax}{arg\,max}

\usepackage{amsmath}

\title{Named Entity and Relation Extraction with Multi-Modal Retrieval}

\author{Xinyu Wang$^{\diamondsuit\ddagger}$, Jiong Cai$^{\diamondsuit\ddagger}$, Yong Jiang$^{\ast}$, Pengjun Xie, 
\textbf{Kewei Tu$^{\diamondsuit}$\thanks{\hspace{1mm} Yong Jiang and Kewei Tu are the corresponding authors. $^{\ddagger}$: equal contributions. This work was done when Xinyu Wang was visiting StatNLP Research Group at SUTD.}, \and Wei Lu$^\clubsuit$} \\
 $^\diamondsuit$School of Information Science and Technology, ShanghaiTech University \\
 Shanghai Engineering Research Center of Intelligent Vision and Imaging \\
 Shanghai Institute of Microsystem and Information Technology, Chinese Academy of Sciences \\
 University of Chinese Academy of Sciences \\
 $^\clubsuit$StatNLP Research Group, Singapore University of Technology and Design \\
  {\tt \{wangxy1,caijiong,tukw\}@shanghaitech.edu.cn} \\
  {\tt \{jiangyong.ml,xpjandy\}@gmail.com} \\
  {\tt luwei@sutd.edu.sg} \\
}

\begin{document}
\maketitle

\begin{abstract}
Multi-modal named entity recognition (NER) and relation extraction (RE) aim to leverage relevant image information to improve the performance of NER and RE. 
Most existing efforts largely focused on directly extracting potentially useful information from images (such as pixel-level features, identified objects, and associated captions).
However, such extraction processes may not be knowledge aware, resulting in information that may not be highly relevant.
In this paper, we propose a novel \underline{M}ulti-m\underline{o}dal \underline{Re}trieval based framework (MoRe).
MoRe contains a text retrieval module and an image-based retrieval module, which retrieve related knowledge of the input text and image in the knowledge corpus respectively.
Next, the retrieval results are sent to the textual and visual models respectively for predictions.
Finally, a Mixture of Experts (MoE) module combines the predictions from the two models to make the final decision.
Our experiments show that both our textual model and visual model can achieve state-of-the-art performance on four multi-modal NER datasets and one multi-modal RE dataset.
With MoE, the model performance can be further improved and our analysis demonstrates the benefits of integrating both textual and visual cues for such tasks.\footnote{Our code is publicly available at \url{https://github.com/modelscope/AdaSeq/tree/master/examples/MoRe}.} 
\end{abstract}

\section{Introduction}

Utilizing images to improve the performance of Named Entity Recognition (NER) and Relation Extraction (RE) has attracted increasing attentions in natural language processing. Image information can be utilized in various domains such as social media \citep{zhang2018adaptive,moon-etal-2018-multimodal,lu-etal-2018-visual,zheng2021mnre}, movie reviews \citep{10.1145/3474085.3475400} and news \citep{wang2022wikidiverse}. Most of the previous approaches utilize images by extracting information such as feature representations \citep{moon-etal-2018-multimodal,yu-etal-2020-improving-multimodal}, object tags \citep{10.1145/3394171.3413650,zheng2021multimodal} and image captions \citep{wang2022ita} to improve the model performance. 
However, most of image feature, object tag and caption extractors are trained based on datasets such as ImageNet \citep{imagenet_cvpr09} and Visual Genome \citep{krishnavisualgenome}, which mainly contain common nouns\footnote{\url{https://visualgenome.org/data_analysis/statistics}} instead of named entities. As a result, the extractors (especially those of object tags and image captions) often output information about common nouns, which may not be helpful for entity-based task models. Recently, pretrained vision-language models \citep{Tan2019LXMERTLC,chen2020uniter,li2020oscar} have improved the performance of cross-modal tasks such as VQA \citep{Agrawal2015VQAVQ}, NLVR \citep{Young2014FromID} and image-text retrieval \citep{Suhr2019ACF} significantly. However, suffering from the same problem, the pretrained vision-language models do not achieve better performance than textual models in multi-modal NER \citep{Sun2021RpBERTAT,wang2022ita}.

Recently, approaches based on text retrieval have shown their effectiveness over question answering \citep{liu2020k,xu2021human,wang-etal-2022-training}, machine translation \citep{gu2018search,zhang-etal-2018-guiding,xu-etal-2020-boosting}, language modeling \citep{guu2020realm,DBLP:journals/corr/abs-2112-04426}, NER \citep{wang-etal-2021-improving,wang-etal-2022-damo,zhang-etal-2022-domain} and entity linking \citep{zhang-etal-2022-iclr,huang-etal-2022-nlpcc}. The approaches use the input texts as the search query to retrieve the related knowledge in the knowledge corpus (KC), which is a key-value structured memory built from the knowledge source. Besides, humans can recognize the entities (such as famous persons and locations) in the image based on their learned knowledge in practice. When they are not sure about the entities in the image, they can even use image-based retrieval in the search engine to get the related knowledge about the image. Inspired by that, for multi-modal NER and RE models, we believe retrieving the related knowledge of the image can be utilized to help the task models to disambiguate the named entities as well. In this paper, we propose \underline{M}ulti-m\underline{o}dal \underline{Re}trieval based framework (MoRe), which explores the knowledge behind the input image and text pairs for multi-modal NER and RE. MoRe retrieves related knowledge for the input text and image using the textual retriever and image retriever respectively. The text retriever retrieves the most related paragraphs in the KC and the image retriever finds the documents containing the most related images. The retrieval results of each modality are sent to the textual and visual models respectively and used for training on NER and RE tasks. After both of the models are trained, the Mixture of Experts (MoE) module is trained to learn how to combine the model predictions from the two models. 

The contributions of MoRe can be summarized in four aspects:
\begin{enumerate}[leftmargin=*]
    \item We propose a simple and effective way to inject knowledge-aware information into multi-modal NER and RE tasks using multi-modal retrieval, which is rarely introduced on multi-modal NER and RE tasks in previous work.
    \item We empirically show that the knowledge from our text retrieval and image-based retrieval modules can significantly improve the performance of multi-modal NER and RE tasks. 
    \item We further propose MoE for multi-modal NER and RE, which can combine the knowledge from the image and text retrieval modules well. We show MoE can further improve the performance and achieve state-of-the-art accuracy.
    \item We conduct detailed analyses that compare the advantage of the text retrieval module and image-based retrieval module. We show the MoE module can correctly take the advantage of the knowledge from each modal.
\end{enumerate}

\begin{figure*}[t!]
	\centering
	\includegraphics[scale=0.30]{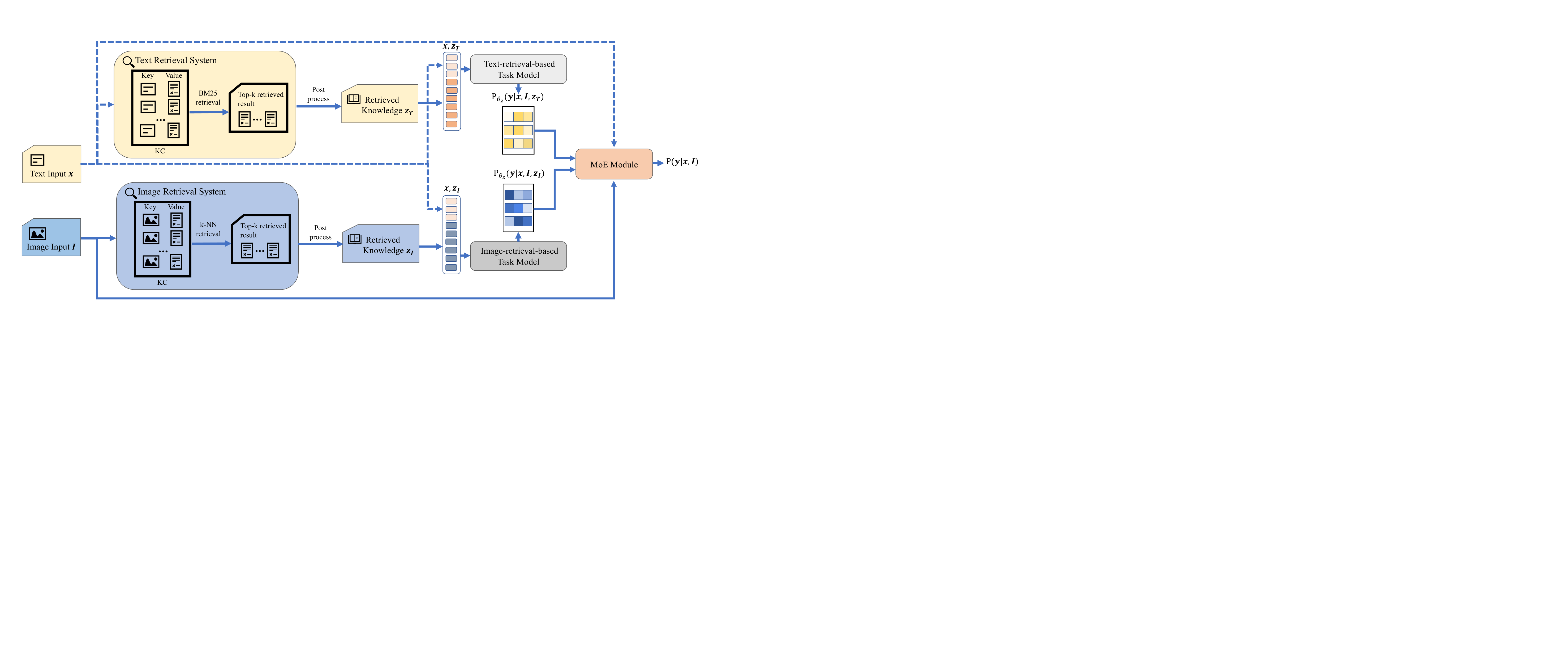}
	\caption{The architecture of MoRe.}
	\label{fig:architecture}
\end{figure*}

\section{MoRe}
Given an input text and image pair $(\vx,\mI)$, where $\vx = \{x_1, \cdots, x_n\}$, MoRe aims to predict the outputs $\vy$. $\vy$ can be the label sequence or the relations for the NER and RE task respectively. MoRe feeds $(\vx,\mI)$ into the multi-modal retrieval module, which returns retrieved texts $\vz_{T}$ and $\vz_{I}$ from a text retrieval module and a image-based retrieval module respectively. The textual task model and visual task model concatenate the input sentence $\vx$ with the retrieved knowledge $\vz_{T}$ and $\vz_{I}$ from KC 
and predict the output distribution $P(\vy|\vx,\mI,\vz_{T})$ and $P(\vy|\vx,\mI,\vz_{I})$ respectively. Finally, the MoE module fuses the predictions from the models based on each modality and makes the final prediction $P(\vy|\vx,\mI)$. The architecture of our framework is shown in Figure \ref{fig:architecture}. 

\subsection{Multi-modal Retrieval Module}
\label{sec:retr}

Text retrieval has been shown to be effective for NER \cite{wang-etal-2021-improving}, as the retrieval result can provide information for disambiguation of complex entities. In order to retrieve related knowledge of the text and image, we design a multi-modal retrieval module. 
We choose Wikipedia, the largest online encyclopedia as our knowledge source because it has a rich collection of articles describing entities and the articles should provide used clues for entity-related tasks. Considering the difficulty of pairing text and image in Wikipedia, we constructed two separate retrieval systems in the module for text and image respectively. 

A retrieval system has two components: KC and knowledge retriever. The KC is denoted by $\{(\vk, \vv)\}$. The knowledge retriever calculates the relevance between an input query $\vq$ and the keys in the KC. The retriever then returns the values corresponding to top-$k$ keys that are most relevant to the query. We denote the retrieved result as $\{ \vt_1, ... \vt_k \}$. A summarization of the two retrieval modules is shown in Table \ref{tab:retrieval}.

\begin{table}[t!]
\small
\setlength\tabcolsep{3pt}
\centering
\begin{tabular}{l|cccc}
\toprule
\textbf{System} & \textbf{Query} & \textbf{Key} & \textbf{Value}  & \textbf{Alg.}    \\
\midrule
Text & $\vx$ & Sentence & Appeared Paragraph & BM25 \\
Image & $\mI$ & Img. Feat. & Introduction Section & $k$-NN \\
\bottomrule
\end{tabular}
\caption{A comparison of text retrieval and image-based retrieval system. Img. Feat.: Image feature, Alg.: Search algorithm, $k$-NN: $k$-nearest-neighbors.}
\label{tab:retrieval}
\end{table}

\paragraph{Textual Retrieval System} 
We retrieve the related knowledge of input text $\vx$ with the textual retrieval system. We build the KC from the articles in Wikipedia. In the KC, each key is the sentence in Wikipedia and the corresponding value is the paragraph where the sentence appears. Considering the retrieval efficiency over a KC with 200 million entries, we choose to use a term-based text retriever. The retriever use BM25 \cite{robertson1995okapi} to calculate the relevance score between key $\vk$ and query $\vx$. 

\paragraph{Image-base Retrieval System} 
According to the style manual of Wikipedia\footnote{\url{https://en.wikipedia.org/wiki/Wikipedia:Manual_of_Style}}, the introduction section of an article is the summary of the most important contents and an image in an article is an important illustrative aid to understanding. Given an input image $\mI$, we search the related images in Wikipedia to collect the knowledge of related entities. Each key in KC is an image from a Wikipedia article and the corresponding value is the concatenation of the article title and introduction section of this article. To find the related images, we use an image encoder to encode the images into feature vectors. For each query $\mI$, it retrieves its $k$-nearest-neighbors with the inner-product metric.

\paragraph{Context Processing}
The top-$k$ results from the KC are concatenated into $\vz = \{\text{[X]},\vt_1, \cdots, \vt_k\}$, where $\text{[X]}$ is a special mark indicating the following sequence to be the retrieved texts. Given a certain transformer-based embedding model, we chunk the retrieved knowledge $\vz$ so that the total subtoken number of $\vx$ and $\vz$ does not exceed the embedding's subtoken number limits.
Since the retrieved texts are usually very long, it is hard to combine the retrieval results from two modalities together as a single $\vz$. As a result, we feed the retrieved knowledge $\vz_T$ and $\vz_I$ to separate models to let the model fully utilize each kind of information.

\subsection{Task Model}
Given the retrieved knowledge $\vz$ (which is either $\vz_{T}$ or $\vz_{I}$), the task model predicts the probability distribution $P(\vy|\vx,\mI,\vz)$ of the task. 

\paragraph{Named Entity Recognition} We take the NER task as a sequence labeling problem, which predicts a label sequence $\vy= \{y_1, \cdots, y_n\}$ at each position. 
The NER model feeds the concatenated text $[\vx;\vz]$ into the transformer-based encoder and gets the token representations $\{\vr_1, \cdots, \vr_n\}$ corresponding to $\vx$:
\begin{displaymath}
\{\vr_1,\cdots,\vr_n,\cdots,\vr_{n+m}\} = \text{embed} ([\vx;\vz])
\end{displaymath}
where $m$ is the length of $\vz$. With the attention module in the transformer-based encoder, the token representations $\{\vr_1, \cdots, \vr_n\}$ contain the retrieved knowledge $\vz$. MoRe then feeds the representations into a linear-chain CRF layer to predict the probability distribution $P(\vy|\vx,\mI,\vz)$ of the label sequence:
\begin{align}
    \psi(y', y, \vr_i) &= \exp(\mW_{y}^{T} \vr_i + \vb_{y',y}) \nonumber\\
    P(\vy|\vx,\mI,\vz) &= \frac{\prod\limits_{i=1}^{n} \psi(y_{i-1}, y_i, \vr_i)}{\sum\limits_{\vy' \in \mathcal{Y}(\vx)} \prod\limits_{i=1}^{n} \psi(y'_{i-1}, y'_i, \vr_i)}\nonumber
\end{align}
where $\psi$ is the potential function. In the potential function, $\mW \in \sR^{d\times t}$ and $\vb \in \sR^{t \times t}$ are parameters for calculating emission and transition scores respectively. $d$ is the hidden size of $\vr$ and $t$ is the label set size. $\mathcal{Y}(\vx)$ denotes the set of all possible label sequences given $\vx$.

\paragraph{Relation Extraction} In RE, the model aims to predict a relation label $P(y|\vx,\mI,\vz)$ given the subject entity $\{x_{\text{start}_s}, \cdots, x_{end_s}\}$ and object entity $\{x_{start_o}, \cdots, x_{end_o}\}$.
We follow PURE \citep{zhong-chen-2021-frustratingly}, which adds special markers in the input $\vx$ to indicate the named entities:
\begin{align*}
\vx^{\prime}= \{x_1, \cdots, \text{<S>}, x_{\text{start}_s}, \cdots, x_{end_s}, \text{</S>},\\ \cdots, \text{<O>}, x_{start_o}, \cdots, x_{end_o}, \text{</O>}, x_n\} 
\end{align*}
In the equation, $\text{<S>}$ and $\text{</S>}$ indicate the start and end of the subject entity while $\text{<O>}$ and $\text{</O>}$ indicate the start and end of the object entity. Similar to the NER model, the RE model feeds the concatenated text $[\vx^{\prime};\vz]$ into the transformer based encoder and gets the token representations of $\text{<S>}$ and $\text{<O>}$, denoted by $\vr_{s}$ and $\vr_{o}$ respectively:
\begin{align*}
&\{\vr_1, \cdots, \vr_{s}, \vr_{\text{start}_s},\\
&\cdots, \vr_{o},  \vr_{start_o}, \cdots, \vr_{n+m+4}\} =\text{embed} ([\vx^{\prime};\vz])
\end{align*}
The probability distribution $P(\vy|\vx,\mI,\vz)$ for relation extraction is then given by:
\begin{align}
    \psi^\prime(y, \vr_s, \vr_o) &= \exp(\mM_{y}^{T} [\vr_s;\vr_o] + \vb^{\prime}_{y}) \nonumber\\
    P(y|\vx,\mI,\vz) &= \frac{\psi^\prime(y, \vr_s, \vr_o)}{\sum\limits_{y^{\prime} \in \mathcal{Y}^\prime} \psi^\prime(y^\prime, \vr_s, \vr_o)}\nonumber
\end{align}
where $\mM\in \sR^{2d\times t^\prime}$ and $\vb^{\prime} \in \sR^{t^\prime \times 1}$ are the parameters for RE. $t^\prime$ is the label set size. $\mathcal{Y}^\prime$ denotes the relation label set.

\subsection{Mixture of Experts}
\label{sec:moe}
As we mentioned in Section \ref{sec:retr}, we use a separate retrieval module for each modality. The retrieval scores of retrieved texts from each modal are therefore not comparable. As a result, it is hard to determine a priori which retrieved knowledge is more helpful to the model performance. We use MoE to alleviate this problem. The MoE module aims to fuse the probability distributions from the textual model and visual model to get better model performance. To obtain the overall probability distribution of generating $\vy$, we treat $e$ as a latent variable and calculate the marginal distribution over $e$, which is:
\begin{align*}
    P(\vy|\vx,\mI) &= \sum\limits_{e \in \{T,I\}} P_{\theta_e}(\vy|\vx,\mI,\vz_e)P_{\theta_c}(e|\vx,\mI) 
\end{align*}
where ${\theta_e}$ is the task model parameters trained with the retrieved knowledge $\vz_e$ and ${\theta_c}$ is the model parameters of MoE. To calculate $P_{\theta_c}(e|\vx,\mI)$, we use a text encoder and an image encoder to extract the representation of $\vx$ and $\mI$ respectively:
\begin{align*}
    &\vr_T = \text{TextEncoder}(\vx); \vr_I = \text{ImageEncoder}(\mI)\\
    &P_{\theta_c}(T|\vx,\mI) = \sigmoid(\mU_{y}^{T} [\vr_T;\vr_I] + \vb^{\star})\\
    &P_{\theta_c}(I|\vx,\mI) = 1- P_{\theta_c}(T|\vx,\mI)
\end{align*}
where $\mU\in \sR^{(d_T+d_I)\times t}$ and $\vb^\star \in \sR^{t \times 1}$. $d_T$ and $d_I$ are the dimension of $\vr_T$ and $\vr_I$ respectively. The final prediction from MoE module is given by:
\begin{align*}
    \hat{\vy} & = \argmax_{\vy \in \hat{\mathcal{Y}}} P(\vy|\vx,\mI) \\
    &= \argmax_{\vy \in \hat{\mathcal{Y}}} \sum\limits_{e \in \{T,I\}} P_{\theta_e}(\vy|\vx,\mI,\vz_e)P_{\theta_c}(e|\vx,\mI) 
\end{align*}
where $\hat{\mathcal{Y}}$ can be $\mathcal{Y}(\vx)$ for NER or $\mathcal{Y}^\prime$ for RE. 
In RE, $\hat{\vy}$ can be easily calculated by finding the largest probability among all possible relation label $y$. However, for NER with a linear-chain CRF layer, $\vy$ represents the corresponding label sequence of the input $\vx$. The possible label sequence set $\mathcal{Y}(\vx)$ is exponential in size. As a result, it is difficult to calculate the weighted summation of two probability distributions (i.e. $P_{\theta_{T}}(\vy|\vx,\mI,\vz_T)$ and $P_{\theta_{I}}(\vy|\vx,\mI,\vz_I)$) with an exponential number of possible label sequence. Instead of directly calculating the equation, we approximate this process by assuming the label at each position can be independently determined:
\begin{align*}
    &P_{\theta_e}(\vy|\vx,\mI) \approx \prod\limits_{i=1}^n P_{\theta_e}(y_i|\vx,\mI)    \\
    &\hat{\vy} {\approx} \argmax\limits_{y_1,\cdots, y_n \in \mathcal{Y}^{\star}} \sum\limits_{e \in \{T,I\}} \prod\limits_{i=1}^n P_{\theta_e}(y_i|\vx,\mI)P_{\theta_c}(e|\vx,\mI)\\
    &\hat{y_i} = \argmax \limits_{y_i \in \mathcal{Y}^{\star}} \sum\limits_{e \in \{T,I\}} P_{\theta_e}(y_i|\vx,\mI)P_{\theta_c}(e|\vx,\mI)
\end{align*}
where $\mathcal{Y}^{\star}$ is the NER label set. We use forward-backward algorithm to calculate the marginalized probability distribution $P_{\theta_e}(y_i|\vx,\mI)$:
\begin{align}
\alpha(y_i)&=\sum\limits_{\{y_0,\dots,y_{i-1}\}} \prod\limits_{k=1}^{i} \psi(y_{k-1}, y_k, \vr_k)\nonumber\\
\beta(y_i) &= \sum\limits_{\{y_{i+1},\dots,y_n\}} \prod\limits_{k=i+1}^{n} \psi(y_{k-1}, y_k, \vr_k)\nonumber\\
P_{\theta_e}(y_i|\vx,\mI)
&\propto \alpha(y_i) \times \beta(y_i) \nonumber 
\end{align}


\subsection{Training}
\paragraph{Named Entity Recognition} We use the negative log-likelihood (NLL) as the training loss for the input sequence with gold labels $\vy^*$:
\begin{align}
\mcL_{\text{NER-NLL}}(\theta_e) = - \log P_{\theta_e}(\vy^*|\vx, \mI, \vz_e) \label{eq:nll_loss}
\end{align}

\paragraph{Relation Extraction} Similar to NER, we calculate the NLL loss with the gold label $y^*$:
\begin{align}
\mcL_{\text{RE-NLL}}(\theta_e) = - \log P_{\theta_e}(y^*|\vx, \mI, \vz_e) \label{eq:re_nll_loss}
\end{align}

\paragraph{Mixture of Experts} Given the trained task models with parameters ${\theta_{T}}$ and ${\theta_{I}}$, the MoE model is trained with NLL loss with $P_{\theta_c}(\vy|\vx,\mI)$. The parameters of trained task model ${\theta_{T}}$ and ${\theta_{I}}$ are fixed during the training of MoE.

\section{Experiments}

\subsection{Settings}
\paragraph{Retrieval System Configuration}
For the retrieval systems, we build the KCs using the English Wikipedia dumps. We convert the dumps into plain text and download the images appearing in the articles. To take advantage of the rich anchors in Wikipedia, we mark them with a special tag. For example, the anchor of \textit{``Alan Turning''} is tagged and the text \textit{``Alan Turing published an article ...''} is transformed into \textit{``<e: Alan\_Turing> Alan Turing </e> published an article ...''}. There are about 200 million entries in the KC for textual retrieval and 4 million entries in the one for image-based retrieval. We build the term-based textual retriever with the search engine ElasticSearch\footnote{\url{https://www.elastic.co/}}. We use ViT-B/32 in CLIP to encode images in the feature-based image retriever and use Faiss \cite{johnson2019billion} for efficient search. For both retrieval modules, we use the top-$10$ retrieval candidates.

\paragraph{Datasets}
For NER, we show the effectiveness of MoRe on Twitter-15, Twitter-17, SNAP and WikiDiverse datasets\footnote{The datasets are available at
\url{https://github.com/jefferyYu/UMT}, \url{https://github.com/Multimodal-NER/RpBERT} and \url{https://github.com/wangxw5/wikiDiverse}.} containing 4,000/1,000/ 3,257, 3,373/723/723, 4,290/1,432/1,459, 6,312/ 755/757 sentences in train/development/test split respectively. The Twitter-15 dataset is from \citet{zhang2018adaptive}. The SNAP dataset is from \citet{lu-etal-2018-visual}. The Twitter-17 dataset is a filtered version of SNAP constructed by \citet{yu-etal-2020-improving-multimodal}. These $3$ datasets are from the social media domain and are the most commonly used datasets for multi-modal NER. The WikiDiverse dataset is a very recent multi-modal entity linking dataset constructed by \citet{wang2022wikidiverse} based on Wikinews. The dataset has annotations of entity spans and entity labels. We convert the multi-modal entity linking dataset into a multi-modal NER dataset to further show the effectiveness of MoRe on the news domain. For RE, we use MNRE dataset\footnote{\url{https://github.com/thecharm/MNRE}}. The dataset is constructed by \citet{zheng2021mnre} based on user tweets and contains 12,247/1,624/1,614 sentences in train/development/test split. MNRE dataset is also from the social media domain.

\paragraph{Model Configuration}
In all experiments of MoRe, we use XLM-RoBERTa large \citep[XLMR;][]{conneau-etal-2020-unsupervised} model as the encoder of task models, which has a strong ability for modeling the contexts. For the text and image encoder in MoE, we use the same CLIP model as it in the retrieval system to extract the text and image features.

\paragraph{Training Configuration}
During training, we finetune the models by AdamW \citep{loshchilov2018decoupled} optimizer. In all experiments, we use the grid search to find the learning rate for the embeddings within $[1\times 10^{-6}, 5\times 10^{-5}]$. We use a learning rate of $5\times 10^{-6}$ and a batch size of $4$ for task model training. Following ITA \citep{wang2022ita}, we use the cross-view alignment loss to minimize the KL divergence between the output distributions of retrieval based input and original input. For MoE, we use the same learning rate and a batch size of $64$ instead. The task models are trained for $10$ epochs and the MoE models are trained for $50$ epochs. All of the results are averaged from $3$ runs with different random seeds.

\subsection{Results}
\begin{table}[t!]
\small
\setlength\tabcolsep{3pt}
\centering
\begin{tabular}{l|ccccc}
\toprule
          & T-15 & T-17 & SNAP  & Wiki & MNRE    \\
\midrule
\citet{10.1145/3394171.3413650} & 72.92 & - & - & - & - \\
\citet{yu-etal-2020-improving-multimodal} & 73.41 & 85.31 & - & - & -\\
\citet{sun-etal-2020-riva} & 73.80 & - & 86.80  & - & -\\ 
\citet{Sun2021RpBERTAT} &74.90 & - & 87.80 & - & - \\
\citet{zhang2021multi} & 74.85 & 85.51 & -  & - & -\\
\citet{zheng2021mnre} & -  & - & - & - & 65.56\\
\citet{zheng2021multimodal} & - & - & - & - & 66.41\\
{\textbf{ITA}} & 78.03 & 89.75 & 90.15 & 76.87 & 66.89 \\
Ours: \textbf{Baseline}  & 77.29     & 88.68     & 89.35 & 76.01      & 65.77 \\
\textbf{MoRe}$_{\text{Text}}$ & 77.91     & 89.50     & 90.09 & 77.97      & 66.62 \\
\textbf{MoRe}$_{\text{Image}}$  & 78.13     & 89.82     & 90.20 & 77.46      & 67.24 \\
\textbf{MoRe}$_{\text{MoE}}$       & \textbf{79.21}     & \textbf{90.67}     & \textbf{91.10} & \textbf{79.33}      & \textbf{68.60} \\
\bottomrule
\end{tabular}
\caption{A comparison of our approaches and state-of-the-art approaches on multi-modal NER and RE. \textbf{T-15}: Twitter-15, \textbf{T-17}: Twitter-17, \textbf{Wiki}: WikiDiverse. The results of ITA on WikiDiverse and MNRE datasets are reproduced by us.}
\label{tab:main}
\end{table}

We compare MoRe with our baseline and previous state-of-the-art approaches on multi-modal NER and RE. Our baseline is the model without any retrieval module and MoE module. To fully show the effectiveness of our approach, we rerun ITA on WikiDiverse for NER and MNRE for RE. ITA is one of the very recent state-of-the-art approaches to the multi-modal NER, which extracts the image captions, object tags and OCR texts in the image to help NER predictions. For MoRe, we also show the model performance only with the text retrieval module\footnote{
The model is similar to the model of \citet{wang-etal-2021-improving}. Our textual retrieval is based on Wikipedia, while the textual retrieval in \citet{wang-etal-2021-improving} is based on Google. Our local retrieval module is much faster and more practical.} and the performance only with the image-based retrieval module to show the strength of the retrieval module. The results in Table \ref{tab:main} show that MoRe outperforms all of the previous state-of-the-art approaches\footnote{Note that all of the previous approaches do not use any retrieval techniques for the tasks.}. Only with the text retrieval module or the image-based retrieval module, our model performance is competitive and even outperforms ITA. On WikiDiverse dataset, our models have more improvements compared with ITA. The possible reason is that our approach can retrieve more helpful information from KC in the news domain while the caption and object extractors do not perform well in this domain. This shows the performance of ITA may be limited for certain task domains. Comparing our models with text retrieval, image-based retrieval and our baseline, our retrieval approaches are significantly stronger than our baseline (with Student's t-test with $p<0.05$). In most of the cases, models with the image-based retrieval module perform better than models with a text retrieval module except on WikiDiverse dataset. The possible reason is the knowledge from the text retrieval is more critical in the news domain.

\section{Analysis}
\begin{table}[t!]
\small
\setlength\tabcolsep{3pt}
\centering
\begin{tabular}{l|ccccc}
\toprule
& T-15 & T-17 & SNAP  & Wiki & MNRE    \\
\midrule
\textbf{Avg. Pool.}$_\text{Text}$       & 78.37     & 89.70     & 90.72 & 78.25      & 66.91 \\
\textbf{Avg. Pool.}$_\text{Image}$      & 78.81     & 89.47     & 90.55 & 78.16      & 68.07 \\
\textbf{Avg. Pool.}$_\text{Text+Image}$ & 79.00     & 89.86     & 91.02 & 78.82      & 68.29 \\
\textbf{MoE}$_\text{Text}$        & 78.62     & 89.86     & 90.80 & 78.45      & 68.22 \\
\textbf{MoE}$_\text{Image}$       & 78.88     & 90.32     & 90.78 & 78.24      & 68.44 \\
\textbf{MoE}$_\text{Text+Image}$            & \textbf{79.21}     & \textbf{90.67}     & \textbf{91.10} & \textbf{79.33}      & \textbf{68.60} \\
\bottomrule
\end{tabular}
\caption{A comparison of MoE in MoRe and other variants of MoE. \textbf{Text}/\textbf{Image}: a mixture of two text retrieval/image-based retrieval based models with two random seeds, \textbf{Text+Image}: a mixture of a text retrieval based model and a image-based retrieval based model.}
\label{tab:moe}
\end{table}

\subsection{Comparison with Other Variants of MoE}
To further show the advantage of our MoE module over text and image models, we compare several variants in Table \ref{tab:moe}. In this analysis, we mix two textual models and two image models with different random seeds for comparison. Firstly, we compare our MoE approach with average pooling, which averages the probability distributions over two models. The results show that our MoE approach outperforms all the average pooling approaches significantly (with student's T test with $p<0.05$) except on the SNAP dataset, which shows the effectiveness of MoE. Comparing among the three average pooling approaches, we can find that averaging the probability distributions over the text and image models performs better than averaging the probability distributions of two models from the same modality. Our MoE is also significantly stronger than the single modality MoE approaches on all the datasets (with $p<0.05$). 
Moreover, we find the relative improvements of MoE depend on each specific dataset. For example, the improvements are relatively smaller in T-15 and SNAP compared with the MoE and average pooling while the improvements on T-17 and Wiki datasets are relatively larger. A possible reason is that the importance of text retrieval and image retrieval is almost equal in most of the samples in the T-15 and SNAP datasets. Similarly, the advantage of multi-modal MoE over single-modal MoE depends on the dataset as well. When the retrieved knowledge from images and that from texts are more complementary, the relative improvements will be much higher (e.g. Wiki). 

\begin{table*}[t!]
\small
\setlength\tabcolsep{5pt}
\centering
\begin{tabular}{l|cccc|cccc|cccc}
\toprule
 & \multicolumn{4}{c|}{\textbf{SNAP}} & \multicolumn{4}{c|}{\textbf{WikiDiverse}} & \multicolumn{4}{c}{\textbf{MNRE}}\\
  & LOC   & ORG   & OTHER  & PER   & LOC   & ORG   & OTHER & PER   & LOC   & ORG   & OTHER & PER   \\
\midrule
\textbf{Baseline} & 86.45 & 89.71 & 76.82 & 93.70 & 78.16 & 76.63 & 62.40 & 89.42 & 71.83 & 62.70 & 61.92 & 62.94 \\
\textbf{MoRe}$_{\text{Text}}$  & \textbf{87.90} & 89.85 & \textbf{79.65} & 93.79 & \textbf{79.97} & 76.64 & \textbf{66.32} & 90.68 & 71.85 & 64.67 & \textbf{66.37} & 63.85 \\
\textbf{MoRe}$_{\text{Image}}$ & 87.72 & \textbf{89.88} & 79.00 & \textbf{94.34} & 78.30 & \textbf{78.53} & 64.88 & \textbf{91.41} & \textbf{74.74} & \textbf{65.86} & 65.38 & \textbf{64.24} \\

\bottomrule
\end{tabular}
\caption{Label-wise F1 score on SNAP, WikiDiverse and MNRE datasets.}
\label{tab:labelf1}
\end{table*}

\subsection{How Text Retrieval and Image-based Retrieval Affect Model Prediction}
To further show the advantage of text retrieval and image-based retrieval module in MoRe, we compare the label-wise F1 in Table \ref{tab:labelf1}. For the diversity of domains and tasks, we choose SNAP, WikiDiverse and MNRE as the representative datasets to show the label-wise F1 score. In the WikiDiverse dataset, there are $13$ entity types. We select location, organization, others and person as the representative labels, which are the most common labels in the dataset and are consistent with the entity label set in SNAP and MNRE datasets. For MNRE, we calculate the entity-label-based F1 score, which calculates the relation F1 score for each entity type. For example, if a relation of two entities is predicted as ``/per/org/member\_of'' (which means the subject is person type, the object is organization type and the relationship between them is ``member\_of''), the relation will be counted into the relation F1 score for both ``per'' and ``org'' entities. We use this way to calculate the relation F1 score to analyze how the retrieval system affects each entity label in RE. From the results in Table \ref{tab:labelf1}, we can observe that 1) the models with the retrieval module outperform our baselines over all the labels; 2) the image-based retrieval module in MoRe is much helpful for recognizing person and organization entities; 3) the text retrieval module in MoRe is helpful for recognizing other entities; 4) for location entities, the text retrieval module has an advantage in NER while the image-based retrieval module has an advantage in RE. The possible reason is that the image-based retrieval can easily capture the person and organization entities since people and organization usually appear in the image. However, other entities such as the entity name of creative works and festivals are hard to be presented in the images. The related knowledge of such kinds of entities can be easily found through text retrieval. 

\begin{table}[t!]
\small
\setlength\tabcolsep{4pt}
\centering
\begin{tabular}{l|ccccc}
\toprule
& T-15 & T-17 & SNAP  & Wiki & MNRE    \\
\midrule
\textbf{Baseline}      & 77.29 & 88.68 & 89.35 & 76.01 & 65.77 \\
\textbf{Random}$_{\text{Text}}$  & 77.03 & 88.53 & 89.21 & 75.88 & 65.34 \\
\textbf{Random}$_{\text{Image}}$ & 77.27 & 88.61 & 89.29 & 75.96 & 65.28 \\
\bottomrule
\end{tabular}
\caption{A comparison of our baseline and the model with random retrieval results.}
\label{tab:random}
\end{table}

\subsection{How the Knowledge Quality Affects Performance}
We analyze how the task model will perform when the quality of retrieved knowledge drops. We randomly select the retrieved knowledge from the KCs of text retrieval module and image-based retrieval module respectively and train the models based on the random knowledge. The results in Table \ref{tab:random} shows that in both of the conditions, the model performance drops moderately compared with our baseline that is trained without retrieval results. The observation shows that using the random retrieved knowledge can introduce noises to the model. The improvements of the task models come from the related knowledge provided by our designed retrieval module rather than the extended input sequence length.

\begin{table}[t!]
\small
\centering
\begin{tabular}{l|cccc}
\toprule
& T-15 & T-17 & SNAP  & Wiki\\
\midrule
\textbf{MoRe}$_{\text{Text}}$      & 77.91 & 89.50 & 90.09 & 77.97 \\
\textbf{MoRe}$_{\text{Text-Marg.}}$ & 77.85 & 89.40 & 90.00 & 77.92 \\
\textbf{MoRe}$_{\text{Image}}$       & 78.13 & 89.82 & 90.20 & 77.46 \\
\textbf{MoRe}$_{\text{Image-Marg.}}$  & 78.07 & 89.73 & 90.15 & 77.39 \\
\bottomrule
\end{tabular}
\caption{A comparison of the performance of the MoRe NER models and the performance of their marginal distributions (labeled with Marg.).}
\label{tab:marginal}
\end{table}

\subsection{How Approximation Affects Performance}
In Section \ref{sec:moe}, we propose to calculate the marginal probability distribution of the CRF layer for NER to approximately calculate the MoE target function. To show how the approximation may affect the model performance, we compute the prediction of our NER model by calculating $\argmax\limits_{y_i \in \mathcal{Y}^{\star}} P_{\theta_e}(y_i|\vx,\mI)$ at each position. The results of our task models with the text retrieval and image-based retrieval modules are shown in Table \ref{tab:marginal}. The results show that the approximation only drops the model performance by no more than $0.1$ F1 score. Therefore we can use the approximated probability distribution to calculate the MoE target function, which can be much easier than the original function for the linear-chain CRF layer.

\begin{table}[t!]
\small
\setlength\tabcolsep{4pt}
\centering
\begin{tabular}{lr}
\toprule
Module &	Sentences/Second\\
\midrule
ITA Feature Extraction & 0.7 \\
CLIP Feature Extraction & 6.4\\
Text Retrieval & 64.6\\
Image Retrieval & 650.1\\
Model Prediction & 8.2 \\
\bottomrule
\end{tabular}
\caption{Speed of ITA feature extraction module and each module of MoRe. Note that the CLIP feature extraction includes the text and image feature extraction.}
\label{tab:speed}
\end{table}

\subsection{Speed Comparison}
In Table \ref{tab:speed}, we compare the speed of each module in MoRe on a single Tesla V100 GPU with 16GB memory with a batch size of $1$. To further show the advantage of MoRe, we also calculate the speed of feature extraction parts (i.e. object and caption extractors based on VinVL \citep{zhang2021vinvl}) of ITA\footnote{The prediction speed of ITA is the same as that of MoRe since the input sequence lengths of the models are similar.}. We can observe that the bottleneck of MoRe is the CLIP feature extraction part, but the speed is much faster than the feature extraction module in ITA. The observation shows the speed advantage of MoRe over ITA.


\begin{figure*}
	\centering
	\includegraphics[scale=0.465]{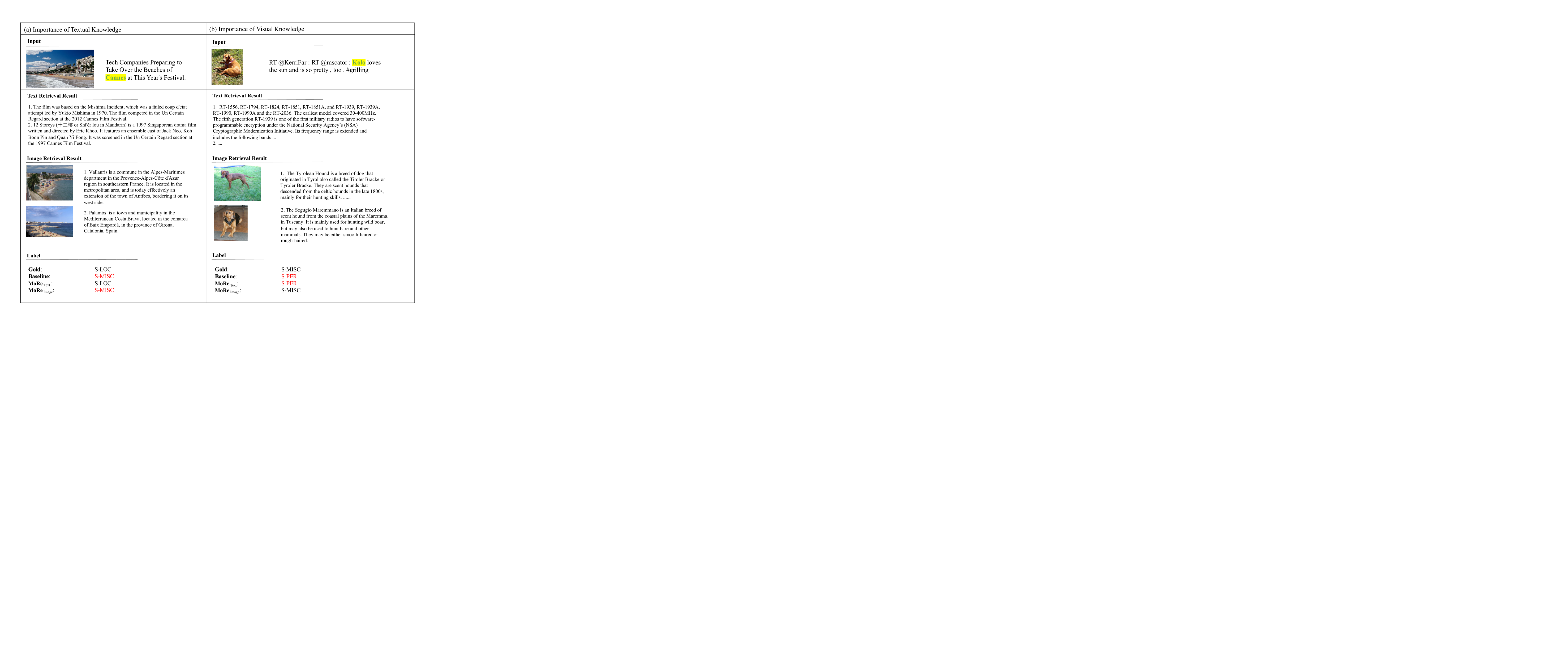}
	\caption{Two case studies of how the text retrieval and image-based retrieval help model predictions.}
	\label{fig:cases2}
\end{figure*}

\subsection{Case Study}
In the case study, we show the importance of knowledge from text retrieval and image-based retrieval. In Figure \ref{fig:cases2} (a), the text input talks about the festival at ``Cannes'' while the image input shows the beaches at the place. The text retrieval results are mainly talking about the Cannes Film Festival while the image-based retrieval results are mainly talking about the similar beaches in the world. For the entity ``Cannes'', the text retrieval results are much more helpful to the disambiguation of the entity since the text retrieval results mention the location ``Cannes'' multiple times. In Figure \ref{fig:cases2} (b), it is hard to recognize the named entity ``Kolo'' is the dog's name only given the input text. The text retrieval also fails to find the related information to the sentence. However, the image-based retrieval returns knowledge about similar kinds of the dog in the input image, which helps the model to recognize the entity ``Kolo'' is possibly the dog's name instead of a person's name.

\section{Related Work}
\paragraph{Introducing Visual Information to Improve Multi-modal NLP Tasks}

In the natural language processing community, improving NLP tasks by introducing visual information becomes a hotspot of recent studies. In many scenarios, there is  a plenty of visual information for a lot of NLP tasks such as NER  \citep{zhang2018adaptive,moon-etal-2018-multimodal,lu-etal-2018-visual}, RE \citep{zheng2021multimodal,zheng2021mnre}, keyphrase prediction \citep{wang-etal-2020-cross-media} and entity linking \citep{10.1145/3474085.3475400,10.1007/978-3-030-73197-7_35,wang2022wikidiverse}. Most of the approaches propose to introduce a special attention mechanism to model the interaction between the representations of objects in the image and the input text \cite{zhang2018adaptive,yu-etal-2020-improving-multimodal,Sun2021RpBERTAT,wang-etal-2020-cross-media,zheng2021multimodal}. \citet{wang-etal-2020-cross-media} and \citet{wang2022ita} additionally introduce OCR texts and image captions to the tasks for further improvements. Recently, \citet{wang2022ita} suggests that the representations of images and texts are trained separately and the representations are not aligned. It is hard for the newly introduced attention mechanism to model the interaction. They
propose to convert an image into the text to ease the alignment problem between text and image. They convert the image into object tags, image captions, and OCR texts for the model. However, the approach may be limited to the training domain of the image information extractor. In comparison, we explore the related knowledge of an image instead of the surface information of an image. The KC can be much easier to build for the specific domains since building it only requires a large scale of domain-specific unlabeled data.

Pretrained vision-language models such as LXMERT \cite{tan-bansal-2019-lxmert}, UNITER \citep{chen2020uniter}, Oscar \citep{li2020oscar}, E2E-VLP \cite{xu-etal-2021-e2e} and mPLUG \cite{li2022mplug} are trained on image-text pairs and achieve significant improvement on tasks like captioning, VQA and image-text retrieval. The pretraining targets at aligning the image and text features into the same space so that the performance of multi-modal tasks can be improved. However, the text representations in pretrained vision-language models are usually not as strong as the pretrained language models. As a result, some of the recent work \cite{Sun2021RpBERTAT,wang2022ita} find that the pretrained vision-language models do not perform well on multi-modal NLP tasks such as NER.

\paragraph{Retrieval-based NLP}
For knowledge-intensive NLP tasks, retrieval is an effective methods to utilize external knowledge. The knowledge retrieval has been applied to a lot of NLP tasks such as question answering \cite{liu2020k,karpukhin-etal-2020-dense, NEURIPS2020_6b493230,xu2021human, izacard-grave-2021-leveraging}, machine translation \cite{gu2018search,zhang-etal-2018-guiding,xu-etal-2020-boosting}, NER \cite{wang-etal-2021-improving,wang-etal-2022-damo,zhang-etal-2022-domain} and entity linking \cite{zhang-etal-2022-iclr,huang-etal-2022-nlpcc}. Compared with these work, our work novelly introduces an image-based retrieval module, which retrieves the knowledge behind the image to improve multi-modal NER and RE tasks. Recently, some of the work introduces the knowledge retrieval to language model pretraining. REALM \cite{guu2020realm} trains the latent knowledge-retriever and knowledge-augmented encoder in an end-to-end manner during the pretraining and finetuning. The generative process in REALM is decomposed into retrieving and predicting. The retrieved knowledge is treated as a latent variable and marginalized. Inspired by the generative process, our MoE module treats whether the retrieved knowledge is from text or image as the latent variable. While REALM aggregates the top-$k$ retrieved knowledge from text with the latent variable, we use it to aggregate the knowledge retrieved from different modalities. Since the retriever is trainable, REALM needs to asynchronously re-embedding and re-indexing all documents during the training. In order to scale with a larger database size, RETRO \cite{DBLP:journals/corr/abs-2112-04426} freezes the retriever and applies a chunked cross-attention mechanism to make use of databases of trillion tokens. For efficiency consideration, we also freeze the retriever module in MoRe. 

\section{Conclusion}
In this paper, we introduce a novel \underline{M}ulti-m\underline{o}dal \underline{Re}trieval based framework that utilizes the knowledge behind the multi-modal inputs. MoRe first retrieves related knowledge of input text and image from a text retrieval module and an image-based retrieval module. MoRe then feeds the retrieved knowledge from the text retrieval module and the image-based retrieval module into the textual and visual task models respectively to make predictions. Given the predictions from the task models of each modality, MoRe combines the prediction by a Mixture of Experts (MoE) module. The MoE module takes the features of each input text and image into consideration and makes the final decision. In our experiments, we show that both our textual model and visual model can achieve state-of-the-art performance on four multi-modal NER datasets and one multi-modal RE dataset. With MoE, the model performance can be further improved. In analysis, we demonstrate the advantage of integrating both textual and visual cues for such tasks over different types of labels.

\section*{Limitations}
In this paper, MoRe requires a textual and a visual KC for the task. We build the KCs based on Wikipedia. However, in some of the scenarios, the KC needs domain-specific unlabeled data for these scenarios. In these cases, the unlabeled data, especially the data with images, should be collected with effort. Moreover, the input length of MoRe is significantly longer than the original input texts since the new inputs contain the retrieved knowledge. As a result, the inference speed should be significantly slower than the speed with the original input texts. Therefore, MoRe may not satisfy some of the time-critical scenarios. However, we can use the techniques such as knowledge distillation \citep{44873} to distill the knowledge from MoRe to smaller models for faster model speed.

\section*{Ethics Statement}
In this paper, we use the publicly available datasets for experiments. For the KCs, we build them based on Wikipedia, which is one of the largest online encyclopedia and is publicly available. Therefore, we believe we do not use any personal data that invades users' privacy.

\section*{Acknowledgements}
This work was supported by the National Natural Science Foundation of China (61976139).

\bibliography{aaai22,acl2021,anthology2,custom}
\bibliographystyle{acl_natbib}


\end{document}